\documentclass{article}
\usepackage{spconf,amsmath,amssymb}
\usepackage{graphicx}
\usepackage{hyperref}
\usepackage{url}
\usepackage{algorithmic}
\usepackage{array}
\usepackage{caption}
\usepackage{subcaption}
\usepackage{setspace}
\usepackage{xcolor}
\usepackage{soul,color} 
\usepackage{cite}

\title{Alzheimer's Disease Diagnostics by a Deeply Supervised Adaptable 
	3D Convolutional Network}

\name{Ehsan Hosseini-Asl\textsuperscript{1}\sthanks{Corresponding Author:---   Tel: (502) 852 3165, Fax: (502) 852 3940,  E-mail: ehsan.hosseiniasl@louisville.edu}, Georgy~Gimel'farb\textsuperscript{2}, Ayman El-Baz\textsuperscript{3}, for the Alzheimer's Disease Neuroimaging Initiative}

\address{\textsuperscript{1}Electrical and Computer Engineering Department, University of Louisville, Louisville, KY, USA.\\
	\textsuperscript{2}Department of Computer Science, University of Auckland, Auckland, New Zealand.\\
	\textsuperscript{3}Bioengineering Department, University of Louisville, Louisville, KY, USA.
}

\begin{document}
\maketitle

\begin{abstract}
Early diagnosis, playing an important role in preventing progress and treating the Alzheimer's disease (AD), is based on 
classification of features extracted from brain images. The features have to accurately capture main AD-related variations of 
anatomical brain structures, such as, e.g., ventricles size, hippocampus shape, cortical thickness, and brain volume. 
This paper proposes to predict the AD with a deep 3D convolutional neural network (3D-CNN), which can learn generic features 
capturing AD biomarkers and adapt to different domain datasets. The 3D-CNN is built upon a 3D convolutional autoencoder,
which is pre-trained to capture anatomical shape variations in structural brain MRI scans. Fully connected upper layers of
the 3D-CNN are then fine-tuned for each task-specific AD classification. Experiments on 
the \emph{ADNI} MRI dataset with no skull-stripping preprocessing have shown our 3D-CNN outperforms 
several conventional classifiers by accuracy and robustness. Abilities of the 3D-CNN to generalize the  
features learnt and adapt to other domains have been validated on the \emph{CADDementia} dataset. 
\end{abstract}


%

\section{Introduction}

The Alzheimer's disease (AD), a progressive brain disorder and the most common case of dementia in the late life,
causes the death of nerve cells and tissue loss throughout the brain, thus reducing the brain volume dramatically through time and 
affecting most of its 
functions~\cite{mckhann1984clinical}. The estimated number of affected people will double for the next two decades, 
so that one out of 85 persons will have the AD by 2050~\cite{alzheimer20142014}. Because the cost of caring for the AD patients 
is expected to rise dramatically, the necessity of having a computer-aided system for early and accurate AD diagnosis becomes 
critical~\cite{bron2015standardized}. 

This paper focuses on developing an adaptable deep-learning-based system for early diagnosis of the AD. Deep learning 
helps to solve such a complex medical diagnosis problem by leveraging hierarchical 
extraction of input data features to improve classification~\cite{plis2014deep}. Several layers of 
feature extractors are trained to form a model being able to adapt to a new domain
by transferring knowledge between different domains and learning distinctive properties of the new 
data~\cite{chen2015net2net,long2015learning}. It has been shown that trained features turn from generality to specificity 
through layers of a deep network~\cite{yosinski2014transferable}, which relates to transferability of features. 

A robust diagnostics of a particular disease should adapt to various datasets, such as, e.g., MRI scans collected by several patient groups, 
as to diminish discrepancies in data distributions and biases toward specific groups. 
Deep learning aims to decrease the use of domain expert knowledge in designing and extracting most appropriate discriminative 
features~\cite{plis2014deep}.  

Several popular non-invasive neuroimaging tools, such as structural MRI (sMRI), 
functional MRI (fMRI), and positron emission tomography (PET) have been investigated for developing such 
a system~\cite{jack2011introduction,mckhann2011diagnosis}. The latter
extracts features from the available images, and a classifier is trained to distinguish between different groups of 
subjects, e.g., AD, mild cognitive impairment (MCI), and normal control 
(NC) groups~\cite{bron2015standardized,cuingnet2011automatic,falahati2014multivariate,sabuncu2015clinical}. 
The sMRI has been recognized as a promising indicator of the AD progression~\cite{bron2015standardized,jack2013tracking}. 

Comparing to the known diagnostic systems outlines below in Section~\ref{sec:prior}, the proposed system 
employs a deep 3D convolutional neural network (3D-CNN) pretrained by 3D Convolutional Autoencoder (3D-CAE) to learn generic discriminative AD features in the lower layers. This captures characteristic
AD biomarkers and can be easily adapted to datasets collected in different domains. To increase the specificity of features in upper layers of 3D-CNN, the discriminative loss function is enforced on upper layers (deep supervision).~\cite{lee2014deeply}. As shown below,  our
system can circumvent shortcomings of its more conventional counterparts.

\section{Prior work}\label{sec:prior}

Voxel-wise, cortical thickness, and hippocampus shape-volume features of the sMRI are used to diagnose 
the AD~\cite{bron2015standardized}. The voxel-wise features are extracted after co-aligning (registering) all 
the brain image data to associate each brain voxel with a vector (signature) of multiple scalar measurements. 
Kl{\"o}ppel et al.~\cite{kloppel2008automatic} used the gray matter (GM) voxels as features and trained an SVM 
to discriminate between the AD and NC subjects. The brain volume in~\cite{fan2007compare} is segmented to 
GM, white matter (WM), and CSF parts, followed by calculating their voxel-wise densities and associating
each voxel with a vector of GM, WM, and CSF densities for classification.
For extracting cortical thickness features, Lerch et al.~\cite{lerch2008automated} segmented the registered brain 
MRI into the GM, WM, and CSF; fitted the GM and WM surfaces using deformable models; deformed and expanded the WM surface 
to the GM-CSF intersection; calculated distances between corresponding points at the WM and GM surfaces to measure the cortical thickness, and used these features for classification. To quantify 
the hippocampus shape for feature extraction, Gerardin et al.~\cite{gerardin2009multidimensional} segmented 
and spatially aligned the hippocampus regions for various subjects and modeled their shape with a series of spherical harmonics. 
Coefficients of the series were then normalized to eliminate rotation--translation effects and used as features for 
training an SVM based classifier. 

Leveraging multi-view MRI, PET, and CSM data for trainable feature extraction of AD prediction involves various 
techniques of machine learning techniques. In particular, Liu et al.~\cite{liu2015inherent} extracted multi-view features using several selected templates in the subjects' MRI dataset. 
Tissue density maps of each template were used then for clustering subjects within each class in order to extract an encoding 
feature of each subject. Finally, an ensemble of support vector machines (SVM) was used to 
classify the subject. Deep networks were also used for diagnozing the AD with different image modalities and clinical data. Gupta et al.~\cite{gupta2013natural} employed 2D Convolutional Neural Network (CNN) for slice-wise feature extraction of MRI scans. To boost the classification performance, CNN was pretrained using Sparse Autoencoder (SAE)~\cite{ng2011ufldl} trained on random patches of natural images.
Suk et al.~\cite{suk2013deep} used a stacked autoencoder to separately extract features from MRI, PET, and cerebrospinal fluid (CSF) 
images; compared combinations of these features with due account of their clinical mini-mental state 
examination (MMSE) and AD assessment scale-cognitive (ADAS-cog) scores, and 
classified the AD on the basis of three selected MRI, PET, and CSF features with a multi-kernel SVM. Later on, a multimodal 
deep Boltzmann machine (BM) was used~\cite{suk2014hierarchical}  to extract one feature from each selected patch of 
the MRI and PET scans and predict the AD with an ensemble of SVMs. Liu et al.~\cite{liu2014multi} 
extracted 83 regions-of-interest (ROI) from the MRI and PET scans and used multimodal fusion to create a set of features to train 
stacked layers of denoising autoencoders. Zhu et al.~\cite{zhu2014novel} proposed a joint regression and prediction model for clinical score and disease group. A linear combination of featurs are used with imposing group lasso constraint to sparsify the feature selection in regression or classification. Li et al.~\cite{li2015robust} developed a multi-task deep learning for both AD classification and 
MMSE and ADAS-cog scoring by multimodal fusion of MRI and PET features into a deep restricted 
BM, which was pre-trained by leveraging the available MMSE and ADAS-cog scores. Zu et al.~\cite{zu2015label} developed a multi-modal classification model, by proposing a multi-task feature selection method. The feature learning method was based on several regression models of different modalities, where label information is used as regularizer ro decrease the discrepancy of similar subjects across different modalities, in the new feature space. Then a multi-kernel SVM is used to fuse modelity-based extracted features for classification. Payan et al.~\cite{payan2015predicting} proposed a 3D Convolutional Neural Network (3D-CNN) for AD diagnosis based on pretraining by SAE. Randomly selected small 3D patches of MRI scans are used to pretrain SAE, where the trained weights of SAE are later used for pretraining of convolutional filters of 3D CNN. Finally, the fully connected layers of 3D-CNN are finetuned for classification.

Comparative evaluations of the sMRI-based feature extraction techniques reveal several limitations for classifying the AD~\cite{bron2015standardized,cuingnet2011automatic,falahati2014multivariate,sabuncu2015clinical}. The voxel-wise feature vectors 
obtained from the brain sMRI are very noisy and can be used for classification only after smoothing and clustering to reduce 
their dimensionality~\cite{fan2007compare}. The cortical thickness and hippocampus model features neglect correlated shape 
variations of the whole brain structure affected by the AD in other ROIs, e.g., the ventricle's volume. Extracted feature vectors highly depend on image preprocessing due to registration errors and noise, so that feature engineering 
requires the domain expert knowledge. Moreover, the developed trainable feature extraction and/or classifiers models~\cite{li2015robust,gupta2013natural,liu2014early,liu2014multi,suk2013deep,suk2014hierarchical,payan2015predicting,liu2015inherent,zhu2014novel,zu2015label} are either dependent on using multi-modal data for feature extraction or biased toward a particular dataset used for 
training and testing (i.e., the classification features extracted at the learning stage are dataset-specific). 

In contrast to all these solutions, the proposed deep 3D-CNN
for learning generic and transferable features 
across different domains is able to detect and extract the characteristic AD biomarkers in one (source) domain and perform task-specific classification in another (target) domain. The proposed network combines a generic feature-extracting stacked 3D-CAE, pre-trained in the source domain, as lower layers with the upper task-specific fully-connected layers, which are fine-tuned in the target 
domain~\cite{yosinski2014transferable,long2015learning}. To overcome the aforementioned feature extraction limitations of the
conventional approaches, the 3D-CAE 
learns and automatically extracts discriminative AD features capturing anatomical variations due to the AD. The pre-trained 
convolutional filters of the 3D-CAE are further adapted to another domain dataset, e.g., to the \emph{ADNI} 
after pre-training on the \emph{CADDementia}. Then the entire 3D-CNN is built by stacking the pre-trained 
3D-CAE encoding layers followed by the fully-connected layers, which are fine-tuned to boost the task-specific classification performance
by employing deep superivision~\cite{lee2014deeply}. Such adaptation of pre-learned generic features to specific tasks~\cite{yosinski2014transferable} allows for
calling the proposed classifier a deeply supervised adaptable 3D-CNN (DSA-3D-CNN).

Data used in the preparation of this article were obtained from the Alzheimer's Disease Neuroimaging Initiative (ADNI) database (adni.loni.usc.edu). The ADNI was launched in 2003 as a public-private partnership, led by Principal Investigator Michael W. Weiner, MD. The primary goal of ADNI has been to test whether serial magnetic resonance imaging (MRI), positron emission tomography (PET), other biological markers, and clinical and neuropsychological assessment can be combined to measure the progression of mild cognitive impairment (MCI) and early Alzheimer?s disease (AD).

\section{Model}
\label{sec:model}
The proposed AD diagnostic framework extracts features of a brain MRI with a source-domain-trained 3D-CAE and performs 
task-specific classification with a deeply supervised target-domain-adaptable 3D-CNN. The 3D-CAE architecture and the AD diagnosis 
framework using the DSA-3D-CNN are detailed in Sections~\ref{sec: 3DCAE} and~\ref{sec:3DCNN},
respectively.
 
\begin{figure*}[hbt!]
\centering
\begin{tabular}{@{}c@{~}c@{}}
\includegraphics[width=0.48\linewidth]{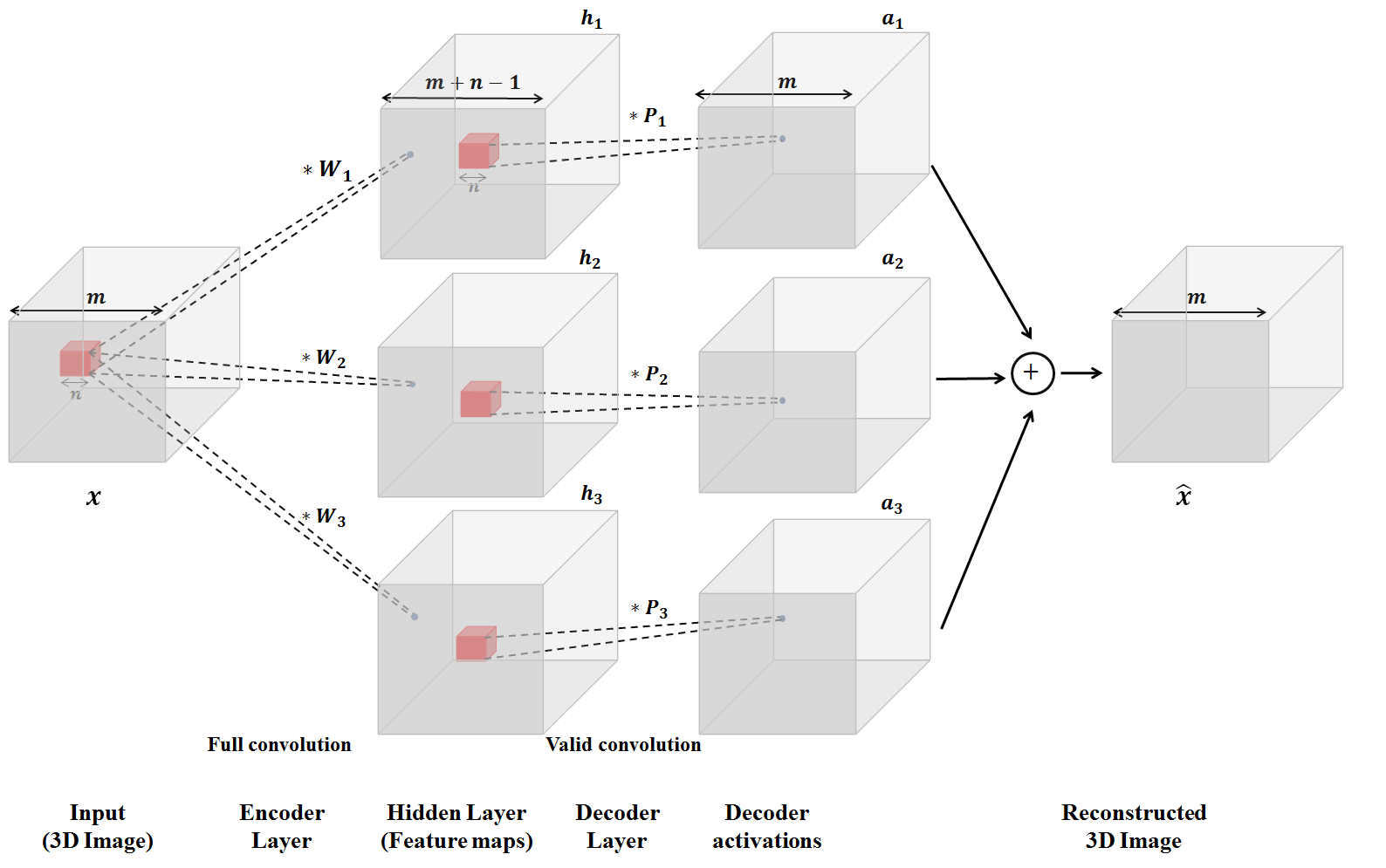} & \includegraphics[width=0.48\linewidth]{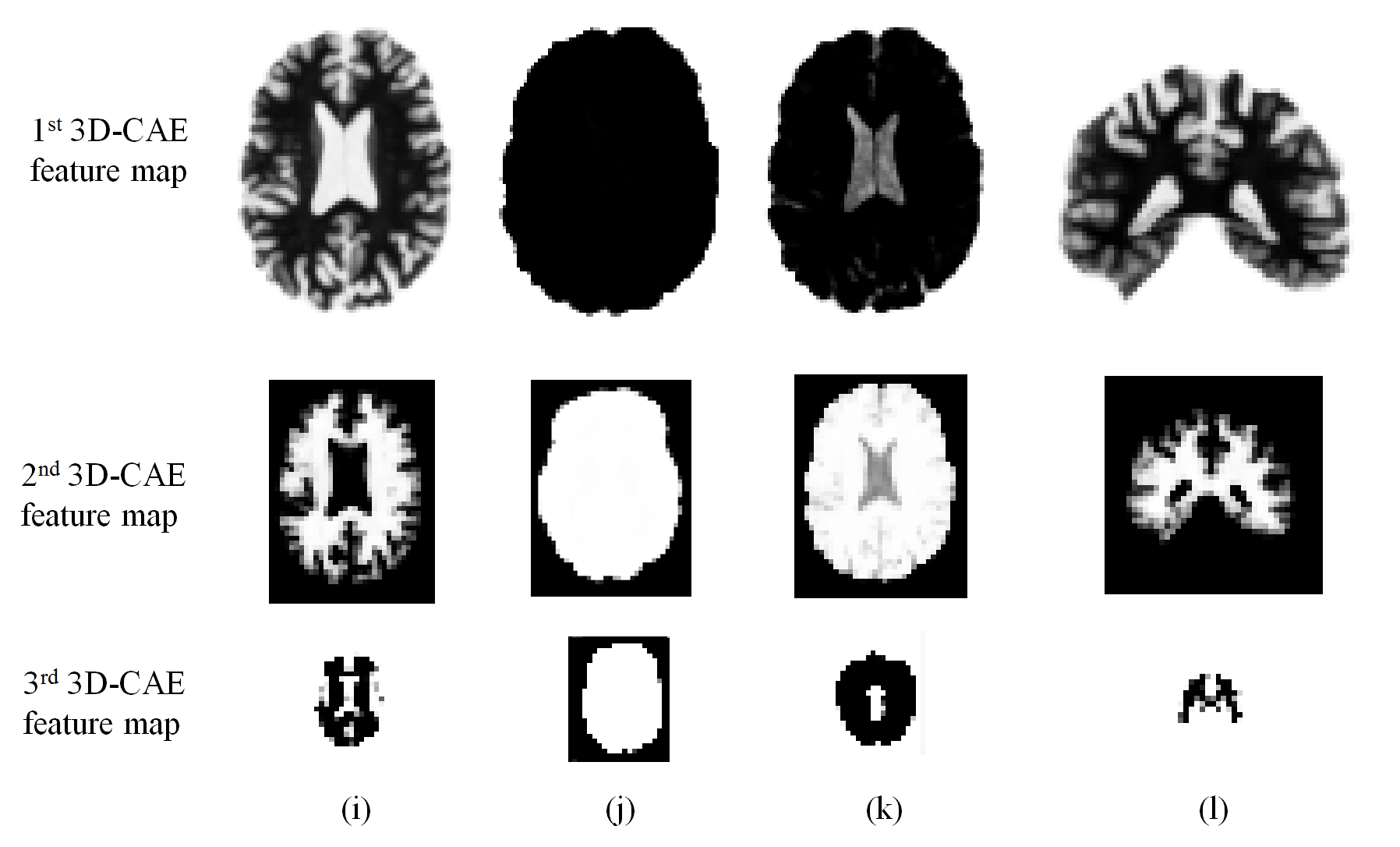}
\end{tabular}
	\caption{Schematic diagram of a 3D-CAE (left) for extracting generic features by convolving and pooling an input 3D
	image (the encoding feature maps are of a larger size, whereas the decoded image keeps the original size), and 
	axial (i,j,k) and sagittal (l) slices (right) of hierarchical 3D feature maps extracted from the source domain, \emph{CADDementia} brain sMRI at three layers of the 
	stacked 3D-CAE: cortex thickness and volume (i), brain size (j), 
	ventricle size (k), and hippocampus model (l). The feature maps are down-sampled at each layer by max-pooling to reduce 
	their size and detect the higher-level features.}\medskip
	\label{fig: 3D-CAE}
	\vspace{-3mm}
\end{figure*}

\subsection{3D Convolutional Autoencoder (3D-CAE)}  
\label{sec: 3DCAE}
Conventional unsupervised autoencoder extracts a few co-aligned scalar feature maps for a set of input 3D images 
with scalar or vectorial voxel-wise signals by combining data encoding and decoding. 
The input image is encoded by mapping each fixed voxel neighborhood to a vectorial feature space in the hidden layer and 
is reconstructed back in the output layer to the original image space. To extract features that capture
characteristic patterns of input data variations, training of the autoencoder employs back-propagation and constraints on 
properties of the feature space to reduce the reconstruction error. 

Extracting global features from 3D images with vectorial voxel-wise signals is computationally expensive and requires too 
large training data sets. This is due to growing fast numbers of parameters to be evaluated in the input (encoding) 
and output (decoding) layers~\cite{lecun1998gradient,ehsan2014autoencoder}. Moreover, although autoencoders with full connections between all nodes of the layers try to 
learn global features, local features are more suitable for extracting patterns from 
high-dimensional images. To overcome this problem, we use a stack of unsupervised CAE with locally 
connected nodes and shared convolutional weights to extract local features from 3D images with possibly long 
voxel-wise signal vectors~\cite{masci2011stacked,makhzani2014winner,Leng20153d}. Each input image is reduced 
hierarchically using the hidden feature (activation) map of each CAE for training the next-layer of CAE.

Our 3D extension of a hierarchical CAE proposed in~\cite{masci2011stacked} is shown in Fig.~\ref{fig: 3D-CAE}. 
To capture characteristic variations of an input 3D image, $\mathbf{x}$, each voxel-wise feature, $h_{i:j:k}$, associated with the 
$i$-th 3D lattice node, $j$-th component of the input voxel-wise signal vector, and $k$-th feature map; $k=[1,\ldots,K]$, 
is extracted by a moving-window convolution 
(denoted below $\ast$) of a fixed $n\times n \times n$ neighborhood, $\mathbf{x}_{i:\mathrm{neib}}$,
of this node with a linear encoding filter specified by its weights, $\mathbf{W}_k=[\mathbf{W}_{j:k}:\:j=1,\ldots,J]$ for each relative
neighboring location with respect to the node $i$ and each voxel-wise signal component $j$, 
followed by feature-specific biases, $\mathbf{b}_k=[b_{j,k}:\:j=1,\ldots,J]$ 
and nonlinear transformations with a certain activation function, $f(\cdot)$: 
\begin{equation}\label{eq:encode}
h_{i:j:k} = f\left(\mathbf{W}_k\ast\mathbf{x}_{i:\mathrm{neib}} + b_{j:k}\right)
\end{equation}
The latter function is selected from a rich set of  constraining differentiable functions, including, in particular, the 
sigmoid, $f(u) = (1+\exp(-u))^{-1}$ and rectified linear unit (ReLU), $f(u)=\max{(0,u)}$~\cite{glorot2010understanding}. Since the
3D image $\mathbf{x}$ in Eq.~(\ref{eq:encode}) has the $J$-vectorial voxel-wise signals, actually, the weights $\mathbf{W}_k$ define
a 3D moving-window filter convolving the union of $J$-dimensional signal spaces for each voxel within the window. 

To simplify notation, let $\mathbf{h}_k=\mathbb{T}\left(\mathbf{x}:\:\mathbf{W}_k,\mathbf{b}_k,f(\cdot)\right)$ 
denote the entire encoding of 
the input 3D image with $J$-vectorial voxel-wise signals with the $k$-th 3D feature map, $\mathbf{h}_k$, such that its 
scalar components are obtained with Eq.~(\ref{eq:encode}) using the weights $\mathbf{W}_k$ and 
bias vectors $\mathbf{b}_k$ for a given voxel neighborhood. 
The similar inverse transformation, $\mathbf{T}_{\mathrm{inv}}(\ldots)$, with the same voxel neighborhood, but 
generally with the different convolutional weights, $\mathbf{P}_k$,  biases, $\mathbf{b}_{\mathrm{inv}:k}$, and, 
possibly, activation function, $g(\cdot)$,
decodes, or reconstructs the initial 3D image:
\begin{equation}\label{eq:decode}
\widehat{\mathbf{x}} = \sum\limits_{k=1}^K 
\underbrace{\mathbf{T}_{\mathrm{inv}}\left(\mathbf{h}_k:\:\mathbf{P}_k,\mathbf{b}_{\mathrm{inv}:k}, g(\cdot)\right)}_{\mathbf{a}_k}
\end{equation}
Given $L$ encoding layers, each layer $l$ generates an output feature image, $\mathbf{h}_{(l)} = [\mathbf{h}_{(l):k}:\:k=1,\ldots,K_l]$,
with $K_l$-vectorial voxel-wise features and receives the preceding output, $\mathbf{h}_{(l-1)} = [\mathbf{h}_{(l-1):k}:\:k=1,\ldots,K_{l-1}]$ 
as the input image (i.e., $\mathbf{h}_{(0)} = \mathbf{x}$).

The 3D-CAE of Eqs.~(\ref{eq:encode}) and~(\ref{eq:decode}) is trained by minimizing the mean squared reconstruction error 
for $T$; $T\ge1$, given training input images,
$\mathbf{x}^{[t]}$; $t=1,\ldots,T$, 
\begin{equation}
E(\boldsymbol{\theta})=\frac{1}{T}\sum_{t=1}^{T}\parallel\widehat{\mathbf{x}}^{[t]}- \mathbf{x}^{[t]}\parallel^{2}_{2}
\label{eq: cost}
\end{equation}
where $\boldsymbol{\theta}=[\mathbf{W}_{k};\mathbf{P}_k;\mathbf{b}_{k};\mathbf{b}_{\mathrm{inv}:k}:\:k=1,\ldots,K]$,
and $\parallel\ldots\parallel^{2}_{2}$ denote all free parameters and
the average vectorial $\ell_{2}$-norm over the $T$ training images, respectively. To reduce the number of free parameters, the decoding weights $\mathbf{P}_{k}$ and encoding weights $\mathbf{W}_{k}$ were tied by flipping 
over all their dimensions as proposed in~\cite{masci2011stacked}. The cost function Eq.~(\ref{eq: cost}) was minimized in the parameter
space by using the stochastic gradient descent search and combined with error back-propagation. 

In order to obtain translational invariance, the feature maps, $\mathbf{h}_{(i)}$, are down-sampled by max-pooling, i.e.,
extracting the maximum value of non-overlapping sub-regions. For entangling shape variations in the higher-level feature maps of 
reduced size, the max-pooling output is used for training the higher layer CAE, as shown in Fig.~\ref{fig: 3D-CAE}(a). 
Stacking the encoding 3D-CAE layers (abbreviated 3D-CAES below) halves the size of the feature map at each level of 
their hierarchy~\cite{masci2011stacked}.

\subsection{Transfer Learning and Domain Adaptation}
\label{sec:transfer}
To achieve good performance, supervised learning of a classifier often requires a large training set of labeled data. If
this set is in principle of a too limited size, additional knowledge from building a similar classifier can be involved 
via so-called transfer learning. In particular, the goal classifier based on a deep CNN might employ initial weights been 
learned for solving similar task~\cite{thrun1996learning,caruana1997multitask,raina2006constructing,baxter1997bayesian}.

We focus on domain adaptation~\cite{bridle1990,ben2010theory,crammer2008learning}, or source-to-target adaptation, when a classifier after training from a source data is adapted to the target data. Unlike usual supervised learning with a classifier 
trained from the scratch by minimizing a total quantitative loss from errors on the training data, the domain adaptation 
minimizes the same loss over the target domain by updating the classifier, which has been trained on the source domain. We leverage the 
unsupervised feature learning to transfer features found in the source domain to the target domain in order to boost the 
predictive performance of the deep CNN models~\cite{mesnil2012unsupervised,glorot2011domain}.

\begin{figure*}[htb!]
	\centering
	\includegraphics[width=18cm]{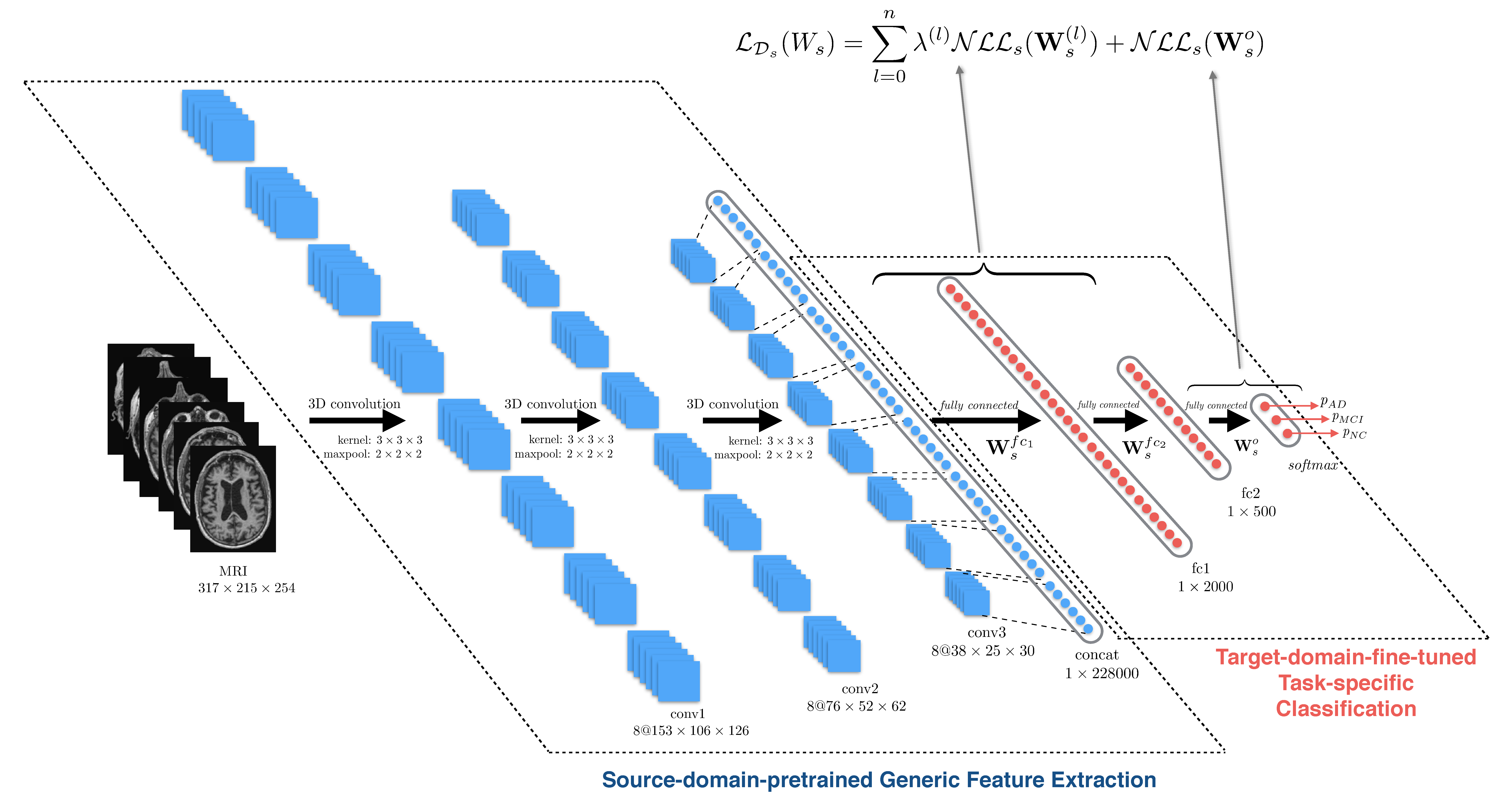}
	\caption{Architecture of the deeply supervised and adaptable 3D CNN (DSA-3D-CNN) for AD diagnosis.}
	\label{fig_model}
\end{figure*}

\subsection{Deeply Supervised Adaptive 3D-CNN (DSA-3D-CNN)}
\label{sec:3DCNN}
While the lower layers of a goal predictive 3D-CNN extract generalized features, the upper layers have to facilitate task-specific 
classification using those features~\cite{long2015learning}. The proposed classifier extracts the generalized features by using
a stack of locally connected lower convolutional layers, while performing task-specific fine-tuning of parameters of the fully connected 
upper layers. Training the proposed hierarchical 3D-CNN consists of pre-training, initial training of the lower convolutional layers,
and final task-specific fine-tuning. At the pre-training stage, the convolutional layers for generic feature extraction are formed as a stack
of 3D-CAEs, which were pre-trained in the source domain. Then these layers are initialized by encoding the 3D-CAE 
weights~\cite{chen2015net2net}, and, finally, the deep-supervision-based~\cite{lee2014deeply} fine-tuning of the upper 
fully-connected layers is performed for each task-specific binary or multi-class sMRI classification.

Due to pre-training on the source domain data, the bottom convolutional layers can extract generic features related to the AD 
biomarkers, such as the ventricular size, hippocampus shape, and cortical thickness, as shown in Fig.~\ref{fig: 3D-CAE}(b). 
We use the Net2Net initialization~\cite{chen2015net2net}, which allows for different convolutional kernel and pool sizes of the 
3D-CNN layers, comparing to those in the 3D-CAE, based on the target-domain image size and imaging specifications, and 
facilitates adapting the 3D-CNN across the different domains. To classify the extracted features in a task-specific way,
weights of the upper fully-connected 3D-CNN  layers are fine-tuned on the target-domain data by minimizing 
a specific loss function. The loss depends explicitly on the weights and is proportional to a negated log-likelihood of the 
true output classes, given the input features extracted from the training target-domain images by the 
pre-trained bottom part of the network. 

\begin{figure*}[htb!]
	\begin{minipage}{0.45\linewidth}
		\centering
		\includegraphics[scale=0.32]{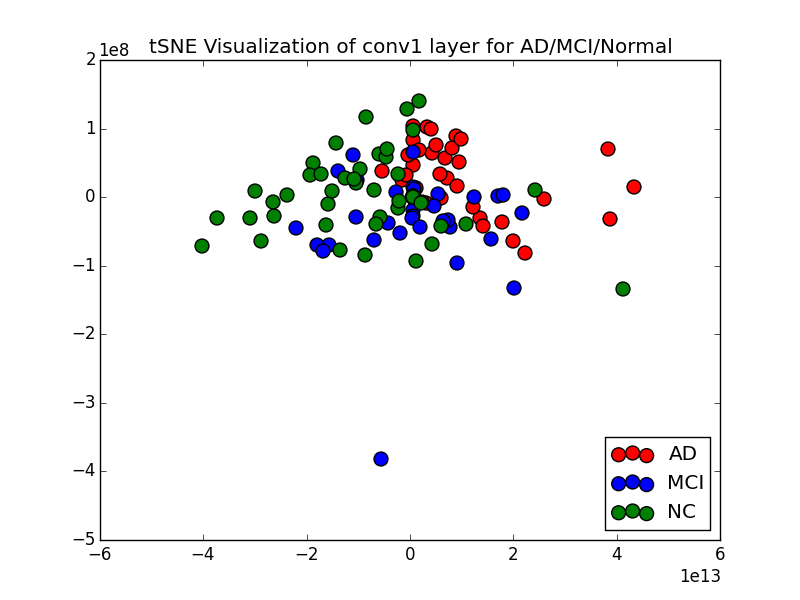}
		
		{{\footnotesize (a) conv1 hidden}}
	\end{minipage}
	\begin{minipage}{0.45\linewidth}
		\centering
		\includegraphics[scale=0.32]{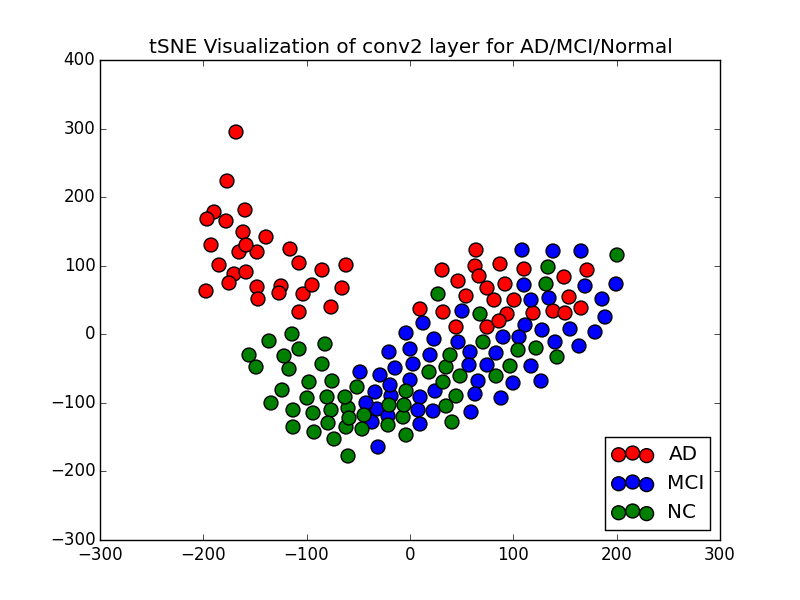}
		
		{{\footnotesize (b) conv2 hidden}}
	\end{minipage}
	
	\begin{minipage}{0.45\linewidth}
		\centering
		\includegraphics[scale=0.32]{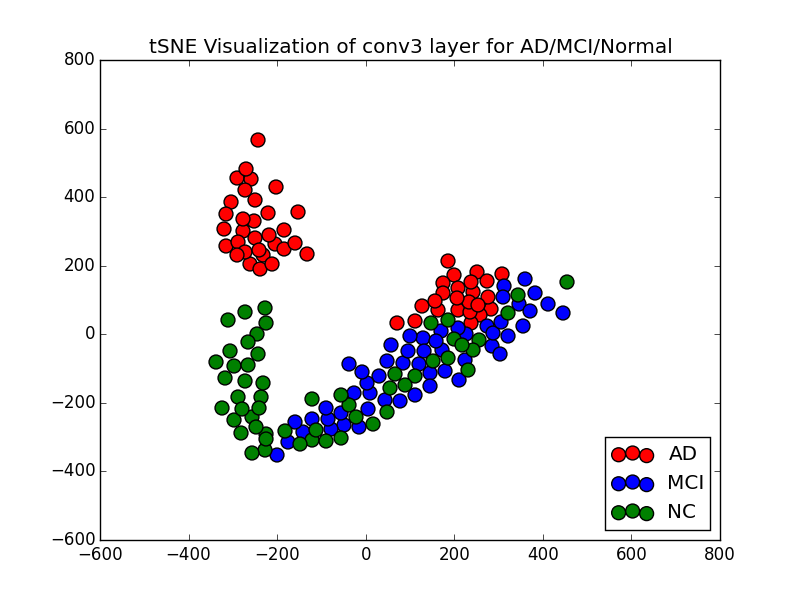}
		
		{{\footnotesize (c) conv3 hidden}}
	\end{minipage}
	\begin{minipage}{0.45\linewidth}
		\centering
		\includegraphics[scale=0.32]{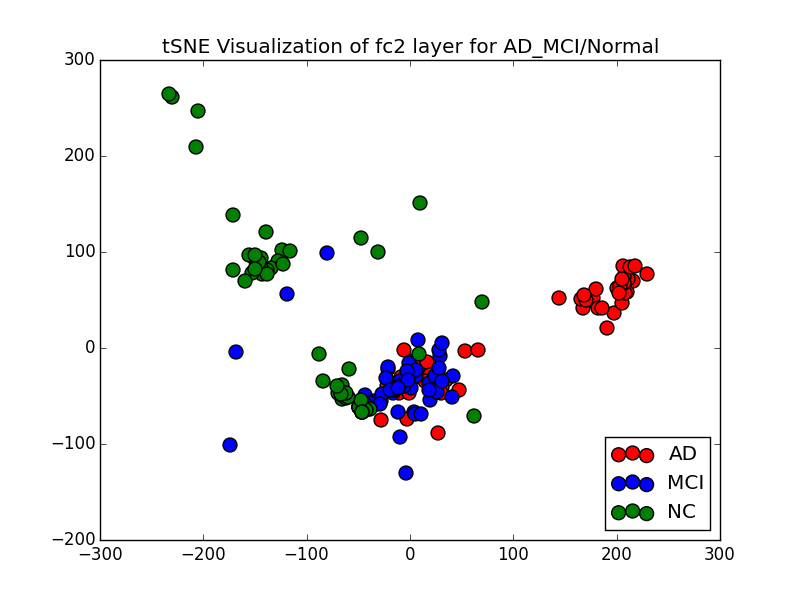}
		
		{{\footnotesize (d) fc2 hidden - AD/MCI/NC}}
	\end{minipage}
	
	\begin{minipage}{0.45\linewidth}
		\centering
		\includegraphics[scale=0.32]{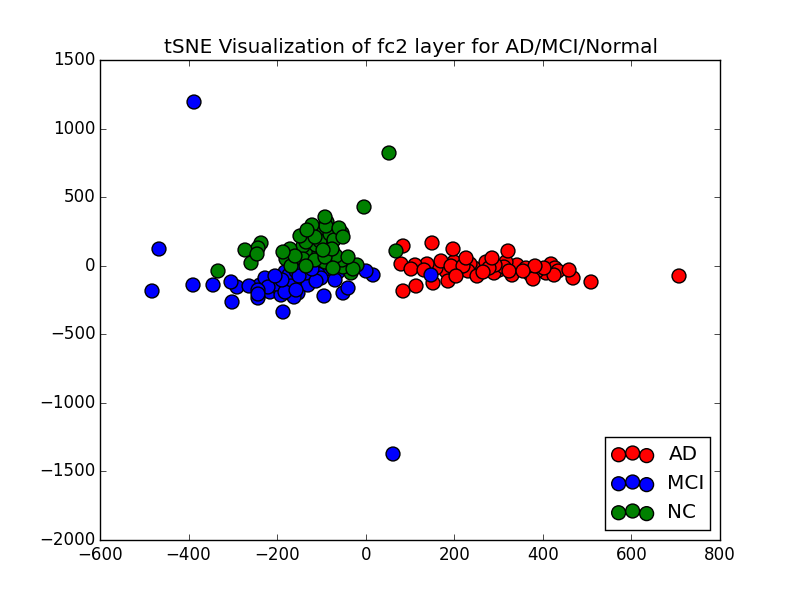}
		
		{{\footnotesize (e) fc2 hidden - AD+MCI/NC}}
	\end{minipage}
	\begin{minipage}{0.45\linewidth}
		\centering
		\includegraphics[scale=0.32]{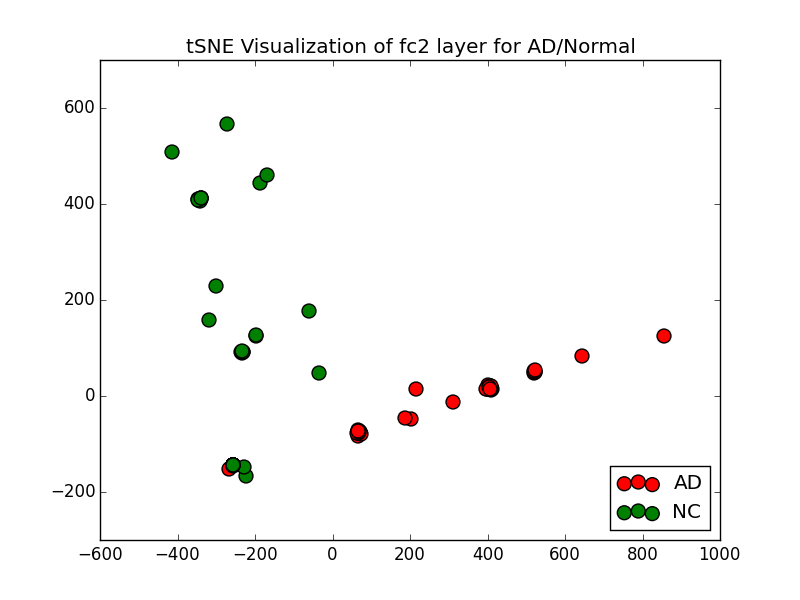}
		
		{{\footnotesize (f) fc2 hidden - AD/NC}}
	\end{minipage}
	
	\begin{minipage}{0.45\linewidth}
		\centering
		\includegraphics[scale=0.32]{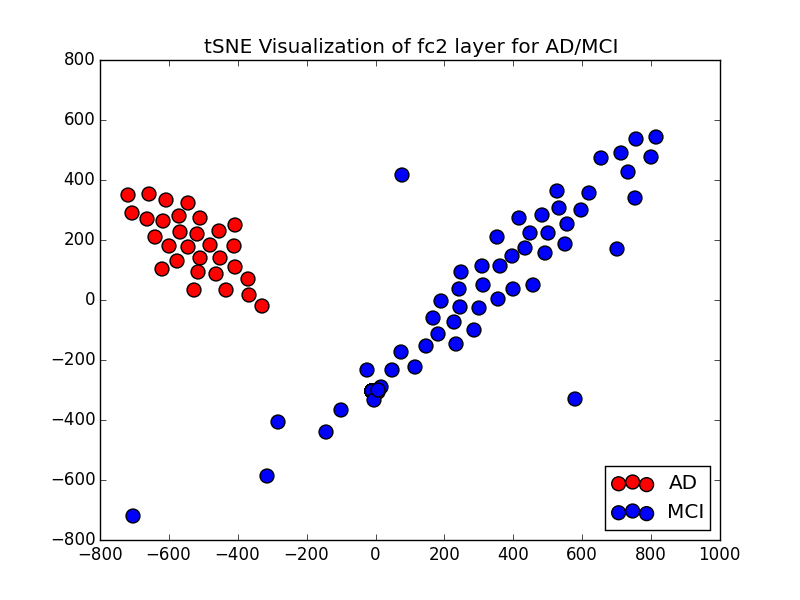}
		
		{{\footnotesize (g) fc2 hidden - AD/MCI}}
	\end{minipage}
	\begin{minipage}{0.45\linewidth}
		\centering
		\includegraphics[scale=0.32]{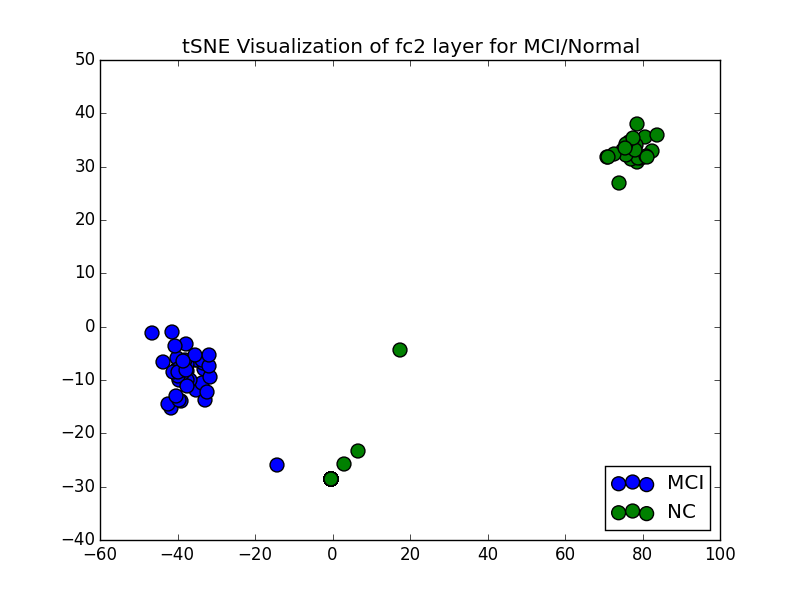}
		
		{{\footnotesize (h) fc2 hidden - MCI/NC}}
	\end{minipage}
\caption{Manifold visualization of target domain (\emph{ADNI}) training data, by t-SNE projection~\cite{van2008visualizing}, in pre-trained generic (a,b,c) 
and fine-tuned task-specific (d,e,f,g,h) DSA-3D-CNN layers.}
	\label{fig:tsne-features}
\end{figure*}

Our implementation of the 3D-CNN uses the ReLU activation functions at each inner layer and
the fully connected upper layers with a softmax top-most output layer (Fig.~\ref{fig_model}), predicting the probability of belonging an
input brain sMRI to the AD, MCI, or NC group. 
The task-specific performance of this 3D-CNN was boosted by deep supervision of its upper 
layers~\cite{lee2014deeply,weston2012deep}. It performs a  task-specific fine-tuning by minimizing  a weighted sum 
of the like log-likelihood-based separate losses, depending each on the weights for each individual fully connected 
upper layer, plus the like loss of the top-most layer doing the softmax transformation of its convolved and biased inputs. 
Discriminative abilities of such a classifier are improved by regularizing the entire hierarchy of
the weights of its hidden layers as shown in Fig.~\ref{fig:tsne-features}. All these properties allow us to call the proposed classifier a 
deeply supervised and adaptable 3D-CNN (DSA-3D-CNN). The Adadelta gradient descent~\cite{zeiler2012adadelta} was used to 
update the pre-trained 3D-CAE and fine-tune the entire DSA-3D-CNN.

\section{Experiments}\label{sec:experiments}
Performance of the proposed DSA-3D-CNN for AD diagnosis was validated on 30 subjects of \emph{CADDementia}, as source domain, and 210 subjects in the 
AlzheimerÕs Disease Neuroimaging Initiative (ADNI) database, as target domain (demographic information mentioned in Table~\ref{tab:demographic}), for five classification tasks: four binary ones 
(AD vs. NC, AD+MCI vs. NC, AD vs. MCI, MCI vs NC) and the ternary classification (AD vs. MCI vs. NC). 
Classification accuracy was evaluated for each test by ten-fold cross-validation. The Theano library~\cite{Bastien-Theano-2012}
was used to develop the deep CNN implemented for our experiments on the Amazon EC2 g2.8xlarge instances 
with GPU GRID K520. 

\begin{table}[htb!]
	\caption{Demographic data for 210 subjects from the target domain, \emph{ADNI} database (STD -- standard deviation).}
		\begin{tabular}{lccc}
			\hline
			Diagnosis & AD & MCI & NC \\
			\hline
			Number of subjects & 70 & 70 & 70 \\
			Male / Female & 36 / 34 & 50 / 20 & 37 / 33 \\
			Age (mean$_{\pm\mathrm{STD}}$) & $75.0_{\pm 7.9}$ & $75.9_{\pm 7.7}$ & $74.6_{\pm 6.1}$ \\
			\hline
		\end{tabular}		
	\label{tab:demographic}
\end{table}

\begin{table*}[htb!]
	\caption{Task-specific performance of the proposed classifier on target domain (\emph{ADNI}) for a selected cross-validation fold 
	(see Section~\ref{sec:exp-classification} and Eq.~(\ref{eq:metrics}).}
	\centering	
		\begin{tabular}{lccccccccccccccc}
			{} & \multicolumn{3}{c}{AD / MCI / NC} & \multicolumn{3}{c}{AD+MCI / NC} & \multicolumn{3}{c}{AD / NC} & 
			\multicolumn{3}{c}{AD / MCI} & \multicolumn{3}{c}{MCI / NC}\\
			{Class} & PPVr & SEN & F1 & PPV & SEN & F1 & PPV & SEN & F1 & PPV & SEN & F1 & PPV & SEN & F1\\
			\hline
			AD & 1.00 & 1.00 & 1.00 & - & - & - & 0.88 & 1.00 & 0.94 & 1.00 & 1.00 & 1.00 &  - & - & -\\
			MCI & 0.60 & 0.80 & 0.69 & - & - & - & - & - & - & 1.00 & 1.00 & 1.00 & 0.92 & 0.97 & 0.94 \\
			AD+MCI & - & - & - & 0.94 & 0.97 & 0.95 & - & - & - & - & - & - & - & - & -\\
			NC & 0.70 & 0.47 & 0.56 & 0.93 & 0.87 & 0.90 & 1.00 & 0.87 & 0.93 & - & - & - & 0.97 & 0.91 & 0.94\\
			\hline
			Mean & 0.77 & 0.76 & 0.75 & 0.93 & 0.93 & 0.93 & 0.94 & 0.93 & 0.93 & 1.00 & 1.00 & 1.00 & 0.95 & 0.94 & 0.94\\
		\end{tabular}
	\label{tab:precision}
\end{table*}

\subsection{Generic and task-specific feature evaluation}
\label{sec:exp-feature}
Special 2D projections of the extracted features in Fig.~\ref{fig: 3D-CAE}(b) illustrate generalization and adaptation abilities of the proposed DSA-3D-CNN (Fig.~\ref{fig_model}). Selected slices of the three feature maps from 
each layer of our stacked 3D-CAE (abbreviated 3D-CAES below) in Fig.~\ref{fig: 3D-CAE}(b), show that the learnt generic convolutional 
filters can really capture features related to the AD biomarkers, e.g., the ventricle size, cortex thickness, and hippocampus model. 
These feature maps were generated by the pre-trained 3D-CAES for the CADementia database. According to these projections, the 
first layer of the 3D-CAES extracts the cortex thickness as a discriminative AD feature of AD, whereas 
the brain size (related to the patient gender), size of ventricles, and hippocampus model are represented by the subsequent layers. 
Each 3D-CAES layer combines the extracted lower-layer feature maps in order to train the higher level for describing more in detail 
the anatomical variations of the brain sMRI . Both the ventricle size and cortex thickness features are combined to extract 
conceptually higher-level features at the next layers. 
Visualized in Fig.~\ref{fig:tsne-features}, projections show capabilities of the extracted higher-layer features to separate the AD, 
MCI, and NC brain sMRI's in the low-dimensional feature space.

Shown in Fig.~\ref{fig:tsne-features} projected manifold distributions of the training ADNI sMRI over the hidden 
layers of our DSA-3D-CNN illustrate discriminative abilities of the generic and task-specific features.
The generic layers (conv1, conv2, and conv3 in Fig.~\ref{fig:tsne-features}(a--c)) gradually enhance the AD, MCI and NC discriminability 
along their hierarchy. The subsequent task-specific classification layers further enhance the discriminability of these 
three ADNI classes, as shown in Fig.~\ref{fig:tsne-features}(d--h). The task-specific features
are highlighted in Fig.~\ref{fig:tsne-features}(d), depicting the distribution of all three classes when the AD+MCI subjects are to be
distinguished from the NC subjects. Obviously, the AD, MCI, and NC cases are projected at closer distances and well separated 
in the task-specific feature space.

\begin{figure*}[htb!]
	\centering
	\includegraphics[width=18cm]{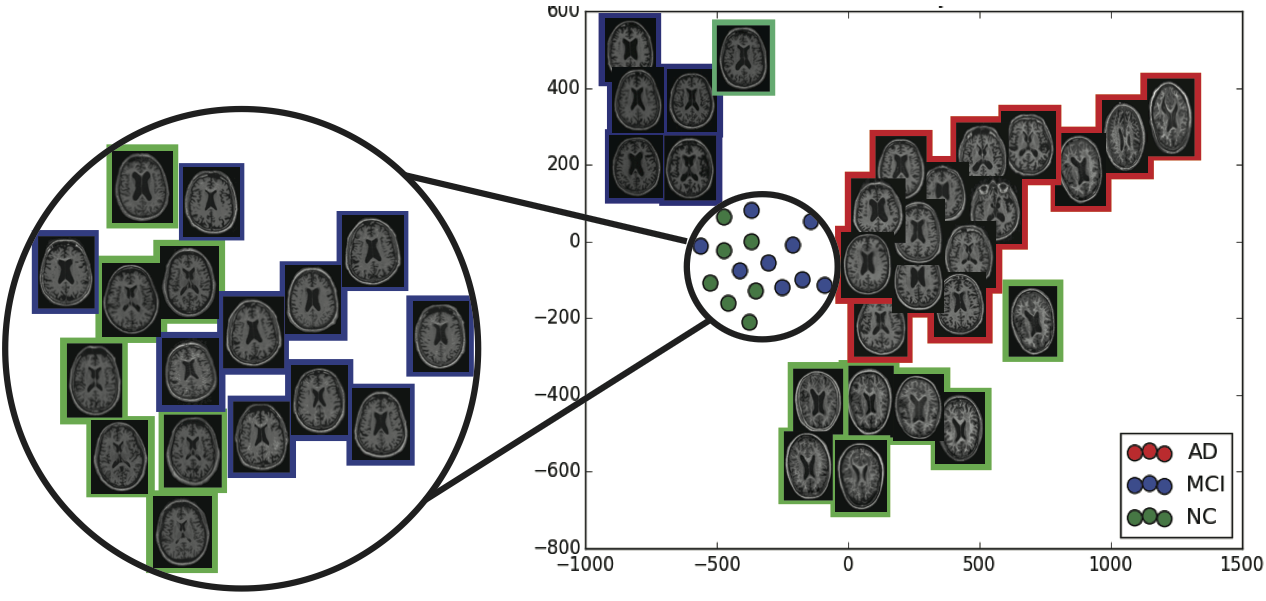}
	\caption{Manifold distribution of target domain (\emph{ADNI}) test data (one fold selected for cross-validation)  in the fc2 layer, visualized by t-SNE projection~\cite{van2008visualizing}.}
	\label{fig:test-tsne}
\end{figure*}

The three-class manifold distribution of the test dataset for ternary (AD vs. MCI vs. NC) classification in Fig.~\ref{fig:test-tsne}
indicates sound discriminative abilities of the learnt features to distinguish between these classes. 
Subject locations in these manifolds indicate that the extracted features correlate with the disease severity. 
The most severe AD cases are at the right-most side of the AD manifold, and the most normal (NC) cases are at 
the bottom of the NC manifold.

\begin{figure*}[htb!]

	\begin{minipage}{0.45\linewidth}
		\centering
		\includegraphics[scale=0.35]{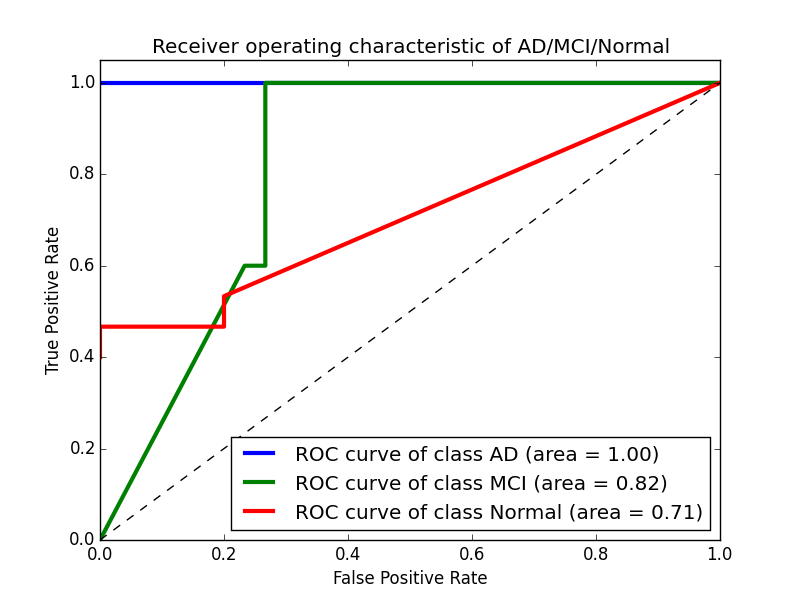}
		%
	\end{minipage}
	\begin{minipage}{0.45\linewidth}
		\centering
		\includegraphics[scale=0.35]{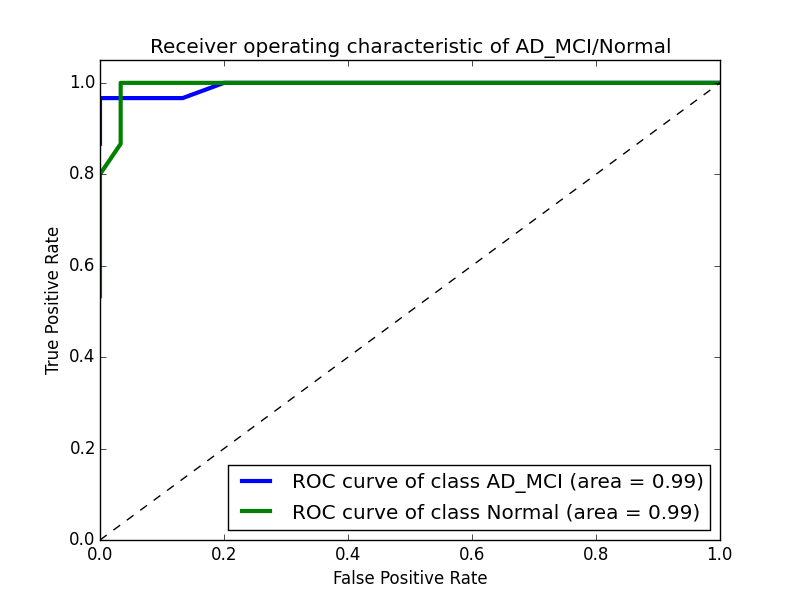}
		%
	\end{minipage}
	
	\begin{minipage}{0.45\linewidth}
		\centering
		\includegraphics[scale=0.35]{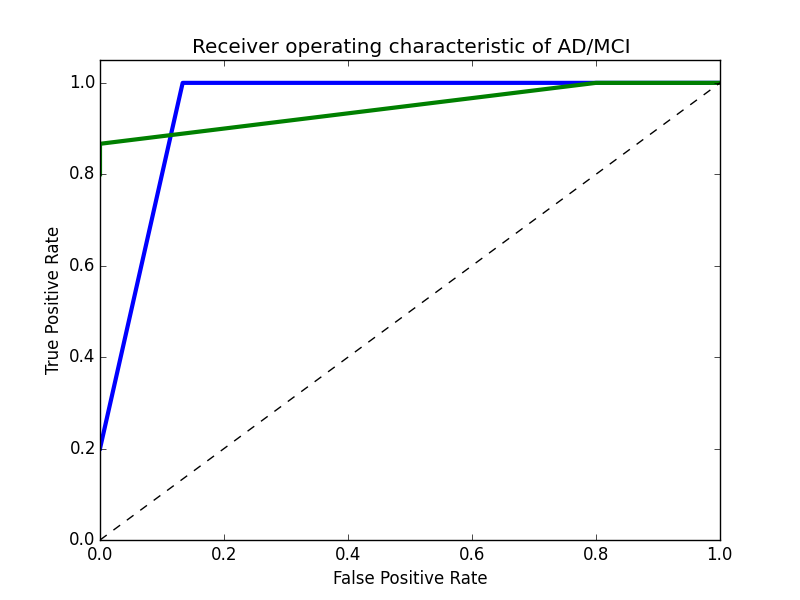}
		%
	\end{minipage}
	\begin{minipage}{0.45\linewidth}
		\centering
		\includegraphics[scale=0.35]{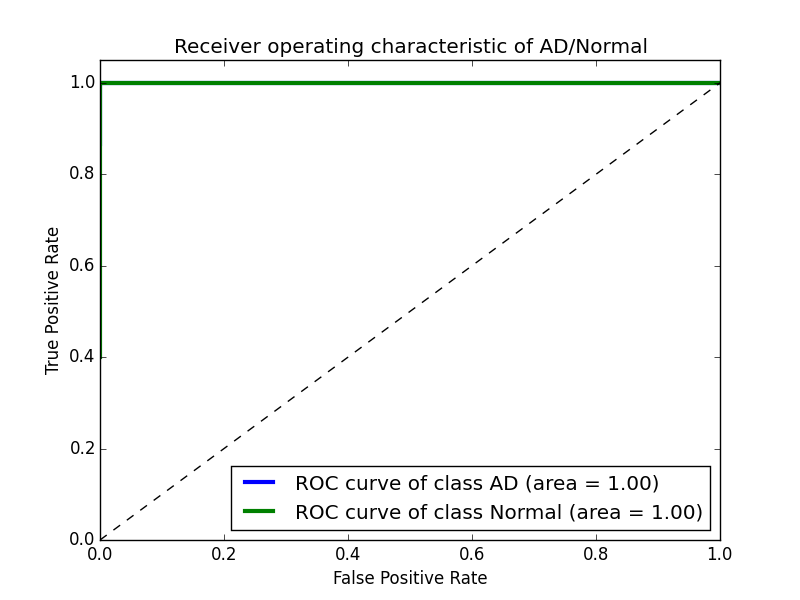}
		%
	\end{minipage}
	
	\begin{minipage}{0.45\linewidth}
		\centering
		\includegraphics[scale=0.35]{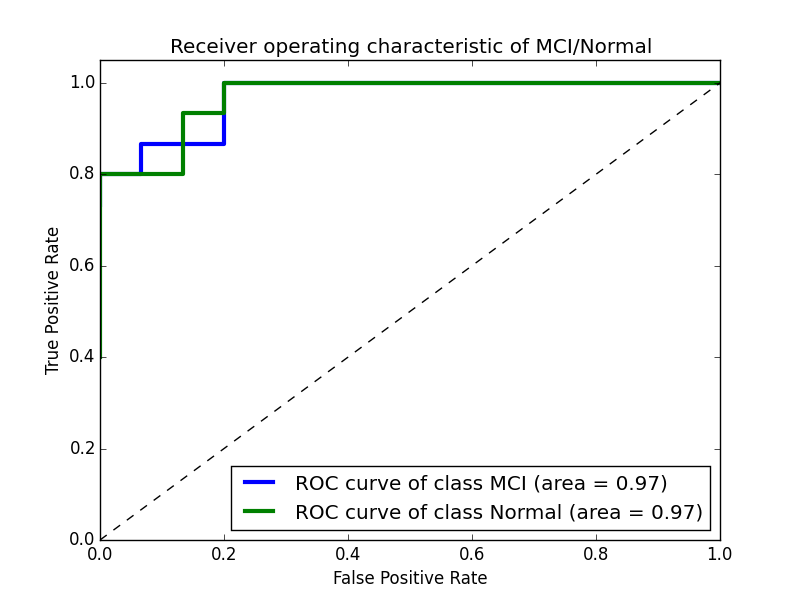}
		%
	\end{minipage}	
	\caption{ROCs and AUC performance scores for the DSA-3D-CNN classifier after fine-tuning to the specific task of 
	distinguishing between (left-to-right) AD / MCI / NC; 
	AD+MCI / NC; AD / NC; AD / MCI, and MCI / NC subjects on target domain (\emph{ADNI}).}
	\label{fig:roc}
\end{figure*}

\begin{table*}[htb!]
	\caption{Performance of the proposed DSA-3D-NCC classifier on target domain (\emph{ADNI}) [$\mathrm{mean}_{\mathrm{STD}}$,\%].}
	\centering
		\begin{tabular}{lcccccccc}
			\hline
Task              & \multicolumn{8}{c}{Performance metrics (Section~\ref{sec:exp-classification}:} \\ \cline{2-9}
                     & ACC & SEN & SPE & BAC & PPV & NPV & AUC & F1-score \\
			\hline
AD / MCI / NC  &  $\bf 94.8_{2.6}$ & $-$                & $-$                & $-$                 & $-$                & $-$                & $-$                & $-$                \\
AD+MCI / NC &  $\bf 95.7_{3.1}$ & $94.8_{4.1}$ & $97.2_{3.8}$ & $96.0_{2.9}$ & $98.4_{2.2}$ & $91.0_{6.8}$ & $96.1_{2.9}$ & $93.9_{4.4}$ \\	AD / NC          & $\bf 99.3_{1.6}$  & $100_0$       & $98.6_{3.1}$ & $99.3_{1.6}$ & $98.6_{3.1}$ & $100_0$        & $99.3_{2.0}$ & $99.4_{1.3}$ \\
AD / MCI         & $\bf 100_0$        & $100_0$       & $100_0$        & $100_0$       & $100_0$        & $100_0$        & $100_0$        & $100_0$       \\
MCI / NC         & $\bf 94.2_{2.0}$ & $97.1_{5.7}$ & $91.4_{4.0}$ & $94.3_{2.0}$ & $91.9_{4.3}$ & $97.1_{4.5}$   & $97.1_{2.0}$  & $94.4_{1.7}$ \\
			\hline
		\end{tabular}
	\label{tab:cnn-perf}
\end{table*}

\begin{table*}[htb!]
	\caption{Comparative performance (ACC,\%) of the classifier vs. seven competitors on \emph{ADNI} dataset (n/a -- non-available).}
	\centering
		\begin{tabular}{llccccc}
			\hline
			{} & {} &\multicolumn{5}{c}{ Task-specific classification [$\mathrm{mean}_{\mathrm{STD}}$,\%].}\\
			\cline{3-7}
			Approach & Modalities & AD/MCI/NC & AD+MCI/NC & AD/NC & AD/MCI & MCI/NC \\
			\hline
Gupta et al.~\cite{gupta2013natural}           & MRI &  $85.0_{\textrm{n/a}}$ & $\textrm{n/a}$ & $94.7_{\textrm{n/a}}$ &  $88.1_{\textrm{n/a}}$ & $86.3_{\textrm{rm}}$ \\
Suk et al.~\cite{suk2013deep}           & PET+MRI+CSF &  $\textrm{n/a}$ & $\textrm{n/a}$ & $95.9_{1.1}$ &  $\textrm{n/a}$ & $85.0_{1.2}$ \\
Suk et al.~\cite{suk2014hierarchical}          & PET+MRI &  $\textrm{n/a}$ & $\textrm{n/a}$ & $95.4_{5.2}$ &  $\textrm{n/a}$ & $85.7_{5.2}$ \\
Zhu et al.~\cite{zhu2014novel}          & PET+MRI+CSF &   $\textrm{n/a}$ & $\textrm{n/a}$ & $95.9_{\textrm{n/a}}$ & $\textrm{n/a}$ & $82.0_{\textrm{n/a}}$ \\
Zu et al.~\cite{zu2015label}                         & PET+MRI &   $\textrm{n/a}$ & $\textrm{n/a}$ & $96.0_{\textrm{n/a}}$ & $\textrm{n/a}$ & $80.3_{\textrm{n/a}}$ \\
Liu et al.~\cite{liu2014multi}                         & PET+MRI & $53.8_{4.8}$ & $\textrm{n/a}$ & $91.4_{5.6}$ & $\textrm{n/a}$ & $82.1_{4.9}$\\
Payan et al.~\cite{payan2015predicting}                             & MRI & $89.4_{\textrm{n/a}}$& $\textrm{n/a}$ & $95.39_{\textrm{n/a}}$ & $86.8_{\textrm{n/a}}$ & $92.1_{\textrm{n/a}}$\\
Liu et al.~\cite{liu2015inherent}                             & MRI & $\textrm{n/a}$& $\textrm{n/a}$ & $93.8_{\textrm{n/a}}$ & $\textrm{n/a}$ & $89.1_{\textrm{n/a}}$\\
Li et al.~\cite{li2015robust}                   & PET+MRI+CSF & $\textrm{n/a}$&  $\textrm{n/a}$ & $91.4_{1.8}$ & $70.1_{2.3}$ & $77.4_{1.7}$\\
Our DSA-3D-CNN 
& MRI & $\mathbf{94.8_{2.6}}$ & $\mathbf{95.7_{3.1}}$ & $\mathbf{99.3_{1.4}}$ & $\mathbf{100_0}$ & $\mathbf{94.2_{2.0}}$ \\
			\hline
		\end{tabular}
	\label{tab:comp}
\end{table*}

\subsection{Classification performance evaluation}
\vspace{-1mm}
\label{sec:exp-classification}
Performance of the proposed DSA-3D-CNN classifier for each specific task listed in Section~\ref{sec:exp-feature} was evaluated 
and compared to competing approaches~\cite{suk2013deep,suk2014hierarchical,zhu2014novel,liu2014multi,liu2015inherent,li2015robust}
by using eight evaluation metrics. Let TP, TN, FP, and FN denote, respectively, numbers of true positive, true negative, false positive, and 
false negative classification results for a given set of data items. Then the performance is measured with the following metrics~\cite{fletcher2012clinical}:  
accuracy (ACC); sensitivity (SEN), or recall; specificity (SPE); balanced accuracy (BAC); positive predictive value (PPV), or precision; 
negative predictive value (NPV), and F1-score, detailed in Eq.~(\ref{eq:metrics}):
\vspace{3mm}
\begin{equation}\label{eq:metrics}
\begin{array}{lll}
\mathrm{ACC}= \frac{\mathrm{TP}+\mathrm{TN}}{\mathrm{TP}+\mathrm{TN}+\mathrm{FP}+\mathrm{FN}}; & & 
\mathrm{F1}    =\frac{ 2\cdot\mathrm{TP} }{ 2\cdot \mathrm{TP}+\mathrm{FP}+\mathrm{FN} }; \\ \\
\mathrm{SEN} = \frac{\mathrm{TP}}{\mathrm{TP}+\mathrm{FN}}; & &
\mathrm{SPE} = \frac{\mathrm{TN}}{\mathrm{TN}+\mathrm{FP}}; \\ \\
\mathrm{PPV} = \frac{\mathrm{TP}}{\mathrm{TP}+\mathrm{FP}};  & &
\mathrm{NPV} = \frac{\mathrm{TN}}{\mathrm{TN}+\mathrm{FN}};  \\ \\
\mathrm{BAC} = \frac{1}{2}( \mathrm{SEN} + \mathrm{SPE}) & &
\end{array} 
\end{equation}
In addition, after building a receiver operating characteristic (ROC) of the classifier, its performance is evaluated by 
the area under the ROC curve (AUC).


Table~\ref{tab:precision} details the class-wise performance of our  DSA-3D-CNN classifier for a selected cross-validation fold and  
five specific classification tasks. The ROCs / AUCs of these tests in Fig.~\ref{fig:roc} and the means and standard deviations of 
all the metrics of Eq.~(\ref{eq:metrics}) in Table~\ref{tab:cnn-perf} indicate high robustness and confidence of the AD predictions by the proposed task-specific DSA-3D-NCC classifier. Its accuracy (ACC) is compared in Table~\ref{tab:comp} with seven 
other known approaches that use either just the same, or even additional inputs (imaging modalities).  
Table~\ref{tab:comp} presents the average results of ten-fold cross-validation of our classifier. Comparing these 
and other aforementioned experiments, the proposed
DSA-3D-CNN outperforms other approaches in all five task-specific cases. This is in spite of employing only a single imaging
modality (sMRI) and performing no prior skull-stripping. 

\section{Conclusion} 
This paper proposed a DSA-3D-CNN classifier which can more accurately predict the AD on structural brain MRI scans than 
several other state-of-the-art predictors. The transfer learning concept is used to enhance generality of the features capturing the AD biomarkers with three stacked 3D CAE network pretrained on \emph{CADDementia} dataset. Subsequently, the features are extracted and used as AD biomarkers detection in lower layers of a 3D CNN network. Then three fully connected layers are stacked on top of the bottom layers to perform AD classification on 210 subjects of ADNI dataset. To boost the classification performance, discriminative loss function was imposed on each fully connected layers, in addition to the output classification layers, to improve the discrimination between subjects. The results demonstrate that hierarchical feature extraction improved in hidden layers of 3D-CNN, by discriminating between AD, MCI, and NC subjects. Seven classification metrics were measured using ten-fold crossvalidation and were compared to the state-of-the-art models. The results have demonstrated the out-performance of the proposed DSA-3D-CNN.

\section{Acknowledgment}
Data collection and sharing for this project was funded by the Alzheimer's Disease Neuroimaging Initiative (ADNI) (National Institutes of Health Grant U01 AG024904) and DOD ADNI (Department of Defense award number W81XWH-12-2-0012). ADNI is funded by the National Institute on Aging, the National Institute of Biomedical Imaging and Bioengineering, and through generous contributions from the following: AbbVie, Alzheimer's Association; Alzheimer's Drug Discovery Foundation; Araclon Biotech; BioClinica, Inc.; Biogen; Bristol-Myers Squibb Company; CereSpir, Inc.; Eisai Inc.; Elan Pharmaceuticals, Inc.; Eli Lilly and Company; EuroImmun; F. Hoffmann-La Roche Ltd and its affiliated company Genentech, Inc.; Fujirebio; GE Healthcare; IXICO Ltd.; Janssen Alzheimer Immunotherapy Research \& Development, LLC.; Johnson \& Johnson Pharmaceutical Research \& Development LLC.; Lumosity; Lundbeck; Merck \& Co., Inc.; Meso Scale Diagnostics, LLC.; NeuroRx Research; Neurotrack Technologies; Novartis Pharmaceuticals Corporation; Pfizer Inc.; Piramal Imaging; Servier; Takeda Pharmaceutical Company; and Transition Therapeutics. The Canadian Institutes of Health Research is providing funds to support ADNI clinical sites in Canada. Private sector contributions are facilitated by the Foundation for the National Institutes of Health (www.fnih.org). The grantee organization is the Northern California Institute for Research and Education, and the study is coordinated by the Alzheimer's Disease Cooperative Study at the University of California, San Diego. ADNI data are disseminated by the Laboratory for Neuro Imaging at the University of Southern California.


\bibliographystyle{IEEEtran}
\bibliography{references}

\end{document}